\documentclass[10pt,twocolumn,letterpaper]{article}

\usepackage[final]{cvpr}      

\definecolor{cvprblue}{rgb}{0.21,0.49,0.74}
\usepackage[pagebackref,breaklinks,colorlinks,allcolors=cvprblue]{hyperref}
\usepackage{tabularx}

\def\paperID{*****} 
\def\confName{CVPR}
\def\confYear{2025}

\title{ An End-to-End Framework for Video Multi-Person Pose Estimation}

\author{Zhihong Wei\\
University of Science and Technology of China\\
weizh588@mail.ustc.edu.cn\\
}

\begin{document}
\maketitle
\begin{abstract}
Video-based human pose estimation models aim to address scenarios that cannot be effectively solved by static image models such as motion blur, out-of-focus and occlusion. Most existing approaches consist of two stages: detecting human instances in each image frame and then using a temporal model for single-person pose estimation. This approach separates the spatial and temporal dimensions and cannot capture the global spatio-temporal context between spatial instances for end-to-end optimization. In addition, it relies on separate detectors and complex post-processing such as RoI cropping and NMS, which reduces the inference efficiency of the video scene. To address the above problems, we propose VEPE (Video End-to-End Pose Estimation), a simple and flexible framework for end-to-end pose estimation in video. The framework utilizes three crucial spatio-temporal Transformer components: the Spatio-Temporal Pose Encoder (STPE), the Spatio-Temporal Deformable Memory Encoder (STDME), and the Spatio-Temporal Pose Decoder (STPD). These components are designed to effectively utilize temporal context for optimizing human body pose estimation. Furthermore, to reduce the mismatch problem during the cross-frame pose query matching process, we propose an instance consistency mechanism, which aims to enhance the consistency and discrepancy of the cross-frame instance query and realize the instance tracking function, which in turn accurately guides the pose query to perform  cross-frame matching. Extensive experiments on the Posetrack dataset show that our approach outperforms most two-stage models and improves inference efficiency by 300\%.

\end{abstract}

\begin{figure}[h]
\centering
\includegraphics[width=0.45\textwidth]{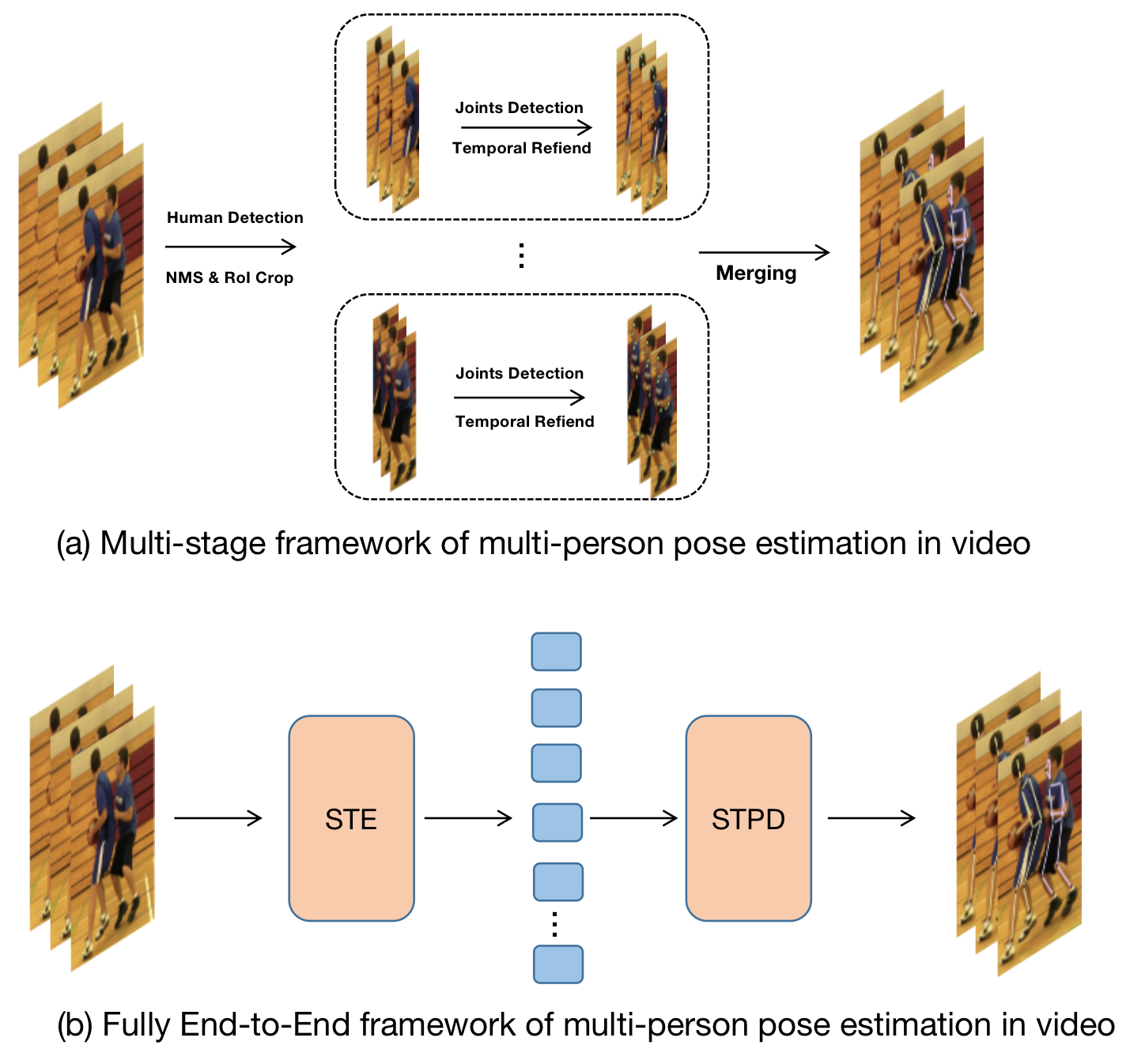} 
\caption{Comparison of multi-stage and end-to-end frameworks. (a) Most of the existing methods are based on this framework: Multi-person pose estimation can be transformed into single-person pose estimation through the use of a human detector and a set of manual components and finally, all human instances are merged.
(b) The end-to-end approach directly views the whole process as a sequence-to-sequence task, where the multi-person pose estimation task is accomplished by feature learning through an encoder and decoding through a decoder.}
\label{fig:cmp}
\end{figure}

\section{Introduction}

Human pose estimation is an important problem in computer vision, which aims to localize the position of keypoints (e.g., knees, ankles, etc.) or body parts. Currently, pose estimation finds extensive applications across various fields, such as action recognition \cite{yan2018spatial}, augmented reality, virtual humans, as well as surveillance and tracking \cite{luo2018lstm}.Early approaches include techniques based on contouring, template matching, feature point detection, and statistical modeling\cite{wang2013beyond,zhang2009efficient,wang2020combining}, which typically rely on traditional computer graphics and image processing techniques and have limited adaptability to complex backgrounds and multi-pose situations. Recently, with the development of deep learning, a number of works\cite{toshev2014deeppose,wei2016convolutional,fang2017rmpe,newell2016stacked,chu2017multi,liu2019towards,wang2020deep,li2022simcc,jiang2023rtmpose,yang2021transpose,Xiao_2018_ECCV,xu2022vitpose} have achieved excellent performance, advancing the development of human pose estimation.

Unfortunately, existing methods designed for static images fall short when it comes to dynamic video images. These methods lack consideration for cues between video frames and disregard the crucial temporal and geometric consistency among frames. As a result, they fail to establish the dependency of human pose cues in temporal sequences, leading to significant performance degradation in scenarios involving motion blur, video out-of-focus, and pose occlusion. Consequently, effectively harnessing temporal sequence information in videos is essential for advancing human pose estimation algorithms in real-life scenarios.

To tackle this issue, various studies \cite{luo2018lstm, artacho2020unipose, wang2020combining} suggest integrating shared sequence characteristics from adjacent frames (support frames). For instance, \cite{luo2018lstm} utilizes a convolutional LSTM to capture both spatial and temporal features, ultimately predicting the pose sequence within videos.\cite{wang2020combining} proposed a 3DHRNet that utilizes 3D convolution to extract spatio-temporal features of video trajectories to estimate pose sequences. Additionally, some methods utilize optical flow or motion estimation to refine pose estimation for the current keyframe \cite{liu2021deep,pfister2015flowing,song2017thin}. For instance, \cite{pfister2015flowing,song2017thin} compute dense optical flow between frames and employ flow-based motion fields to temporally refine pose heatmaps. It is of concern that, optical flow estimation is computationally expensive and prone to fragility in the presence of severe image quality degradation. In another work\cite{liu2021deep}, pose heatmaps of consecutive frames are aggregated, and motion residuals are modelled to enhance pose estimation for keyframes, leveraging implicit or explicit motion priors. Within these frameworks, attention is typically given to all visual motion cues, which often introduces unnecessary motion details, including irrelevant elements like surrounding people or the background. Moreover, most of the above methods use a two-stage detection strategy, where each human instance is first detected and then the instance region in each image frame is cropped, followed by feeding it into a temporal model for prediction. This approach separates the spatial and temporal dimensions and cannot capture the global spatio-temporal context between spatial instances for end-to-end optimization. Furthermore, the computational cost of the two-stage model increases with the number of instances in the image due to the need for additional human detectors and complex post-processing(such as RoI cropping and NMS), leading to less efficient inference, especially in dense scenes. In recent years, end-to-end models have attracted the attention of a wide range of researchers due to their efficient and concise architectures. Compared with traditional two-stage approaches, end-to-end models regress pixel-level poses directly from images without the need for predefined human detectors and complex artificial components, which significantly improves the inference speed and scalability of the models. A comparison of the multi-stage model and the full end-to-end framework structure is shown in Figure.\ref{fig:cmp}.

In this paper, we extend PETR\cite{shi2022end} for video tasks and propose a simple video-based end-to-end pose estimation framework, which naturally treats pose instance objects in the video as a sequence-to-sequence task, and the whole process can be realized with full end-to-end training. The overall construction adopts the Encoder-Decoder structure, which does not require any hand-designed components and greatly improves inference efficiency. The Decoder decoded pose query contains only human pose instances, which naturally filters out background and other interfering factors, and the use of a Transformer to articulate the multi-frame pose query can be naturally generalized to the global context learning at a distance. Our approach incorporates three essential temporal components: the Spatio-Temporal Pose Encoder (STPE), the Spatio-Temporal Deformable Memory Encoder (STDME), and the Spatio-Temporal Pose Decoder (STPD). The Spatio-Temporal Pose Encoder focuses on establishing correlations between spatial pose queries across frames to learn the temporal context of pose instances. Meanwhile, the Spatio-Temporal Deformable Memory Encoder is responsible for extracting cross-frame spatio-temporal visual features for instances using the multi-scale spatial features generated by the Spatial Encoder. Additionally, the Spatio-Temporal Deformable Memory Encoder plays a crucial role in providing valuable human appearance and location cues to the Spatio-Temporal Decoder. To refine the pose results, the spatio-temporal information obtained from the STPE and STDME is decoded in a multi-level cascade using the Spatio-Temporal Pose Decoder (STPD). This decoder further enhances the accuracy and precision of the pose estimation. Furthermore,
to reduce the mismatch problem during the cross-frame pose query matching process, we propose an instance consistency mechanism, which aims to enhance the consistency and discrepancy of cross-frame instance query,  realize the instance tracking function, and to guide the pose query for cross-frame matching, which improves the accuracy of pose instance identification and tracking of Temporal Pose Encoder.

In summary, our contributions can be summarized as follows:

\begin{itemize}
  \item This paper propose a simple and flexible end-to-end pose estimation framework based on video. The framework employs three key spatio-temporal Transformer components: the Spatio-Temporal Pose Encoder (STPE), Spatio-Temporal Deformable Memory Encoder (STDME), and Spatio-Temporal Pose Decoder (STPD), to efficiently utilize temporal information for optimizing human poses.
  \item To reduce the mismatch problem during the cross-frame pose query matching process, this paper proposes an instance consistency mechanism, which aims to enhance the consistency and discrepancy of the cross-frame instance query and realize the instance tracking function, which in turn guides the pose query to perform accurate cross-frame matching.
  \item Extensive experiments on the Posetrack dataset show that VEPE outperforms most two-stage models and improves inference efficiency by up to 300\%.
\end{itemize}

\section{Related Work}
\subsection{End-to-end human pose estimation model}
Current end-to-end multi-person pose estimation\cite{shi2022end,yang2023explicit,liu2023group} frameworks are built by following the designs of DETR\cite{carion2020end} and its variants \cite{zhu2020deformable}. PETR\cite{shi2022end}views pose estimation as a hierarchical set prediction problem and propose the first fully end-to-end pose estimation framework with the advent of DETR. The process begins by predicting a series of human poses through a pose decoder, after which each pose’s keypoints are further refined via a joint (keypoint) decoder. QueryPose\cite{xiao2022querypose} adopts a method similar to Sparse R-CNN\cite{sun2021sparse}, developing two parallel decoders based on RoIAlign, each dedicated to human detection and pose estimation. ED-POSE\cite{yang2023explicit} reinterprets the multi-person pose estimation task as two distinct processes of explicit box detection.It unifies the global person and local keypoints into the same box representation, and they can be optimized by the consistent regression loss in a fully end-to-end manner. GroupPose\cite{liu2023group} introduces a straightforward approach for multi-person pose estimation in an end-to-end manner. By treating each keypoint as an object, it leverages N × K keypoint queries to predict the N × K keypoint positions. Additionally, the method utilizes N instance queries, with each query representing a pose consisting of K keypoints, to provide scoring for the predicted K-keypoint pose. Subsequently, the traditional self-attention is replaced with two consecutive grouped self-attention for performing intra-instance and cross-instance interactions, respectively.

\subsection{Video-based human pose estimation model}

Current image-based methods struggle to adapt effectively to video streams due to inherent challenges in grasping the temporal dynamics across sequential frames.A direct method for cross-frame modelling and utilizing temporal, context is to use the convolutional LSTM proposed in \cite{luo2018lstm,artacho2020unipose}. One primary limitation of this model is its tendency to misalign features across varying frameworks, which consequently diminishes the efficacy of those supporting frameworks. Several studies \cite{pfister2015flowing,song2017thin} utilize optical flow to incorporate motion priors. Generally, these methods calculate dense optical flow between frames and use these motion cues to enhance the accuracy of the predicted pose heatmaps. Nevertheless, estimating optical flow is resource-intensive and can be sensitive to significant declines in image quality. Another line of research \cite{bertasius2019learning,liu2021deep,wang2020combining} explores implicit motion compensation by applying deformable convolutions or 3D CNNs. For instance, \cite{bertasius2019learning,liu2021deep,liu2022temporal} suggest modeling joint movements at multiple granularities based on heatmap residuals and using techniques like pose resampling or pose warping through deformable convolutions.  According to \cite{liu2021deep}, motion offsets are calculated between the keyframe and additional frames, which serve as a foundation for resampling pose heatmaps across successive frames. In both scenarios, the precision of pose estimation is significantly influenced by the effectiveness of optical flow or motion offset calculations. All of the above methods are based on a two-stage approach that requires pre-training of the human detector. The prediction quality of human instances depends on the human detector. The model cannot be trained end-to-end, in which case the lack of effective supervision of the intermediate feature layer may lead to inaccurate pose estimation or even error propagation. In addition, two-stage-based models require the use of artificial components such as NMS, and ROI components, which will greatly increase the burden of inference efficiency.

\begin{figure*}[h]
\centering
\includegraphics[width=1\textwidth]{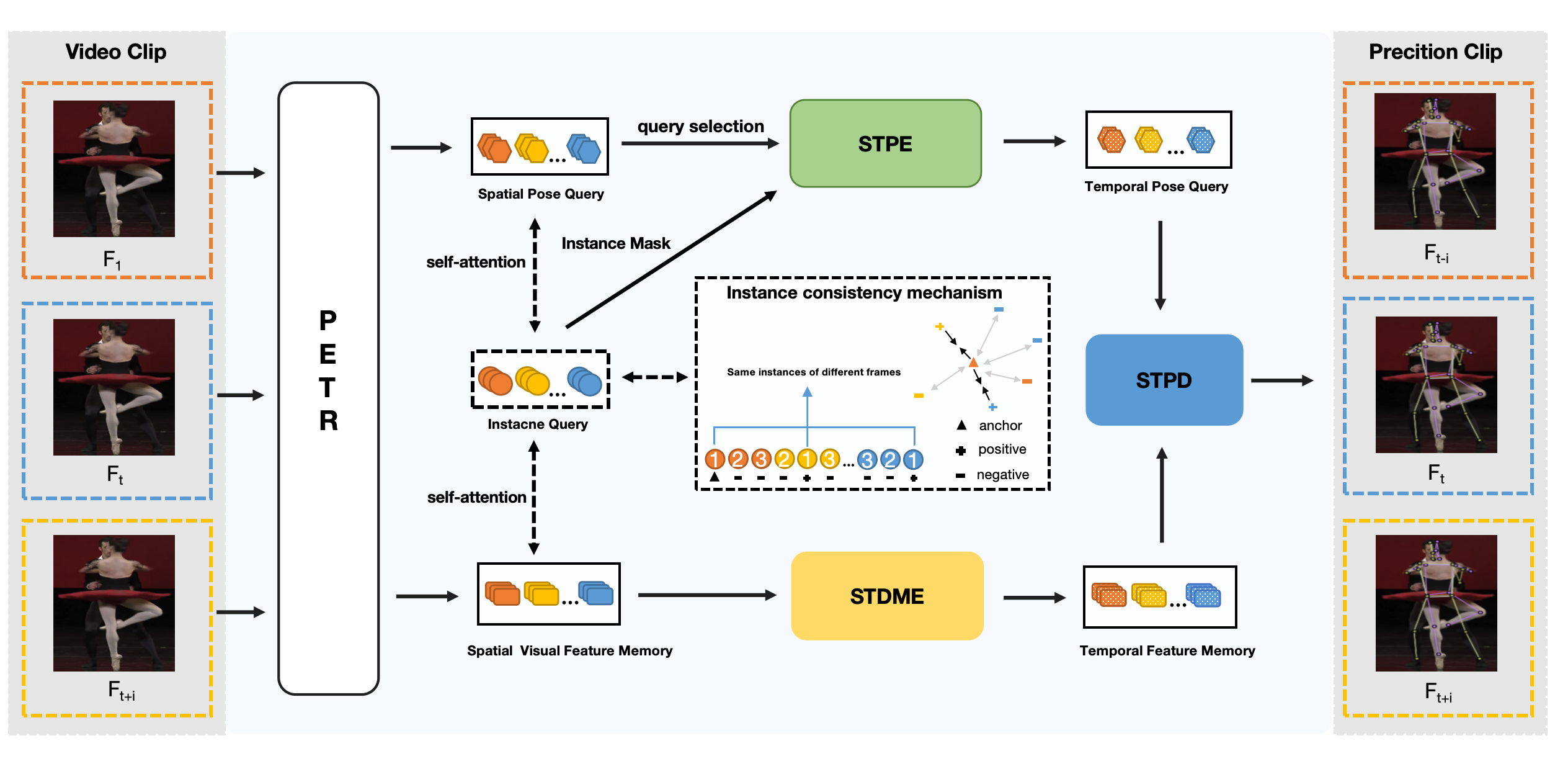} 
\caption{
VEPE model pipeline. The VEPE model pipeline consists of two phases: spatial and temporal. In the spatial phase: spatial visual feature memories and spatial pose queries are obtained by the spatial feature encoder and spatial pose decoder of the PETR model, respectively. In the temporal phase: Initiated with spatial pose queries from PETR first applies query selection to obtain high quality spatial pose queries. Subsequently, high-quality spatio-temporal pose queries are generated through a spatio-temporal learning process with the help of Spatio-Temporal Pose Encoder (STPE) and instance tracking of instance query. Concurrently, the Spatio-Temporal Deformable Memory Encoder (STDME) processes the spatial visual feature memories, temporally encoding them to produce temporal feature memories. Finally, the Spatio-Temporal Pose Decoder (STPD)  outputs the refined pose results by combining the spatio-temporal cues from STPE and STDME. In addition, an instance consistency mechanism is introduced, which aims to optimize the consistency and differentiation of feature representations between instance queries across frames.
}
\label{framework}
\end{figure*}

\section{Method}

\subsection{Review of PETR}
Inspired by the paradigm emerging from object detection \cite{zhu2020deformable}, Shi et al.\cite{shi2022end} have proposed for the first time a fully end-to-end Transformer-based multi-person pose estimation framework, called PETR. The proposed method reduces the problem of pose estimation to a hierarchical ensemble prediction problem by transforming pose estimation into a hierarchical ensemble prediction problem that combines the human body instances and the keypoints in a unified manner. Firstly, a multi-scale feature map is extracted through a mainstream backbone network, and a visual feature encoder takes the spread image features as input and refines them to obtain a refined multi-scale visual feature memory. Next, it operates through multiple randomly initialized trainable pose queries, and the pose decoder is responsible for learning and inferring interactions between objects while evaluating the pose of the instances in the overall image environment. Moreover, the query-based framework is trained utilizing a bipartite matching technique, thereby eliminating heuristic label assignments and removing the necessity for NMS-based post-processing.

\subsection{VEPE Framework}

The overall flow of VEPE is shown in Fig.\ref{framework}. The goal of the framework is to temporally learn static pose results and output fine-tuned, high-quality pose results.

The framework includes the following two parts: the spatial component and the temporal component. The spatial component uses PETR as a spatial feature extractor to extract spatial visual feature memories as well as spatial pose queries through its internal Spatial Encoder and  Decoder. The spatio-temporal components include the Spatio-Temporal Pose Encoder (STPE) for aligning and refining spatial pose queries, the Spatio-Temporal Deformable Memory Encoder (STDME) for fusing spatial visual feature memories, and the Spatio-Temporal Pose Decoder (STPD) for extracting spatio-temporal pose results. In addition, the instance consistency mechanism aims to optimize the consistency and uniqueness of cross-frame instance query feature representations for instance tracking, which ultimately leads to accurate cross-frame matching of pose queries during temporal encoding.


\subsection{Spatio-Temporal Pose Encoder}

The learnable pose query represents a high-level semantic keypoint for capturing human instances. By training this query, we can filter out distractions such as backgrounds, resulting in a query that focuses only on specific instance information. This learnable query facilitates direct temporal interaction with instance objects during cross-frame learning. To facilitate this process, we propose a simple but powerful encoder to facilitate the interaction between cross-frame pose instances. This encoder aligns their feature spaces to convey and interact with information about the same instances in different frames. its idea is to establish correlations between pose queries across the frame space so as to understand the temporal context of the pose instances. We refer to this module as the Spatio-Temporal Pose Encoder (STPE).

\begin{figure}[h]
  \centering
  \includegraphics[width=0.45\textwidth]{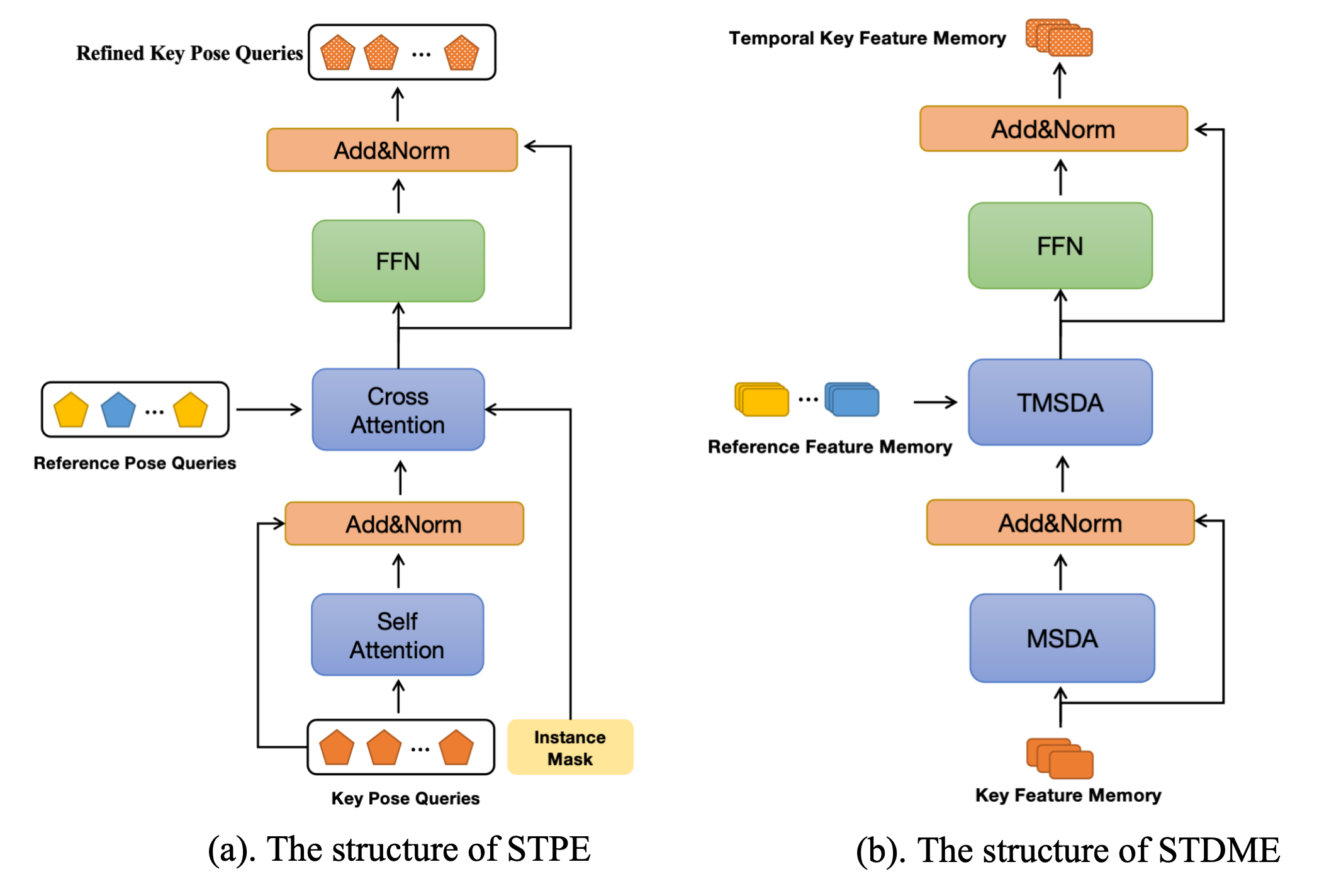}

  \caption{The structure of (a). Spatio-Temporal Pose Encoder (STPE) and (b).Spatio-Temporal Deformable Memory Encoder (STDME). \textbf{STPE}: It takes a keyframe pose query and a reference frame pose query as input and outputs an updated keyframe pose query by aggregating temporal features. \textbf{STDME}: It takes a multi-scale visual feature memory for keyframes and a multi-scale visual feature memory for reference frames as input and outputs an updated multi-scale visual feature memory through temporal feature aggregation. 
}
  \label{fig:stpe_stdme}
\end{figure}

The design of the Spatio-Temporal Pose Encoder (STPE) module is shown in Figure.\ref{fig:stpe_stdme} (a), where the STPE includes a self-attention layer, a cross-attention layer, and a Feedforward Neural Network (FFN). To simplify the explanation of the process in the figure, we designate one frame as the keyframe and the remaining frames as reference frames. Specifically, firstly, the pose query of the keyframe learns the relationship between the instances in the frame through the self-attention layer, and then interacts with the spatial pose object query extracted from the reference frame through the cross-attention layer to compute the common attention between the reference query and the features of the key query, and gradually extracts the useful information to be aggregated to the pose query of the keyframe. In the cross-attention mechanism, it's noteworthy that we use an instance mask to isolate interference from unrelated instances to ensure accuracy. This instance mask is derived from similarity calculations among instance queries, effectively filtering out the most matched targets for each instance in every frame. This process ensures that our algorithm is not impacted by irrelevant instances, which helps reduce unnecessary noise and enhances the overall accuracy and reliability of the model, particularly in complex and highly dynamic scenarios.

In dense scenes, where human bodies often share similar static features, pose queries tend to exhibit significant overlap and resemblance. To tackle this redundancy and leverage the spatio-temporal pose encoder (STPE) for effective learning, we've introduced the Pose Query Selection (PQS) technique. In PQS, we employ a trained instance pose evaluator during the spatial phase to streamline the process. By aggregating spatial pose queries from all frames, the evaluator can then select high-quality pose queries based on specific thresholds. This approach ensures that only the most informative and non-redundant queries are retained, reducing the impact of low-quality pose instances and lowering computational complexity.

\subsection{Spatio-Temporal Deformable Memory Encoder}

The purpose of the Spatio-Temporal Deformable Memory Encoder (STDME) is to encode spatio-temporal features from the multi-scale spatial feature memory of the Spatial Encoder (SE) and provide human appearance and position cues to the Spatio-Temporal Pose Decoder. The ultimate goal is to temporally aggregate visual cues of human pose across frames. Using a plain Transformer encoder\cite{vaswani2017attention} directly can lead to excessive computations due to the similarity of neighbouring features containing similar appearance and background information. Additionally, irrelevant information like the human body background can disrupt the capture of spatio-temporal information. In contrast, deformable attention\cite{zhu2020deformable} samples only a portion of the information based on the learned offset field. The key idea is to focus on a small portion of the surrounding key sampling points and compute similarity for interaction, reducing the computational load of attentional similarity computation and mitigating interference from redundant information.

Multi-scale visual features play an important role in performing human pose estimation tasks. Using multi-scale visual features, pose information can be captured at different scales, including whole-body poses and localized poses, which can provide rich contextual information and help to understand the relationships among human body parts. Therefore, by utilizing extended temporal multi-scale deformable attention (TMSDA), we can link these multi-scale spatial features in the temporal dimension through this manipulation to provide more cross-frame multi-scale visual information for the visual features of the keyframes, which in turn provides additional visual cues for situations such as motion blur and keypoint occlusion.

To simplify the explanation of the process in the figure, we designate one frame as the keyframe and the remaining frames as reference frames. STDME receives as input the multi-scale feature memories of the reference and keyframes and outputs the multi-scale feature memories of the keyframes of the temporal sequence. The specific flow is represented in Figure.\ref{fig:stpe_stdme} (b). The Multi-Scale Deformable Attention (MSDA) captures the contextual relationships among multi-scale visual feature tokens within frames. The Temporal Multi-Scale Deformable Attention (TMSDA) aligns the multi-scale visual feature memory of keyframes and reference frame pose queries, aggregating the most salient temporal information to the keyframe using a partial sampling strategy.
Where Temporal multi-scale deformable attention(TMSDA) is realized as shown in the following equation:

\begin{equation}
    \begin{split}
        \text{TMSDA}(z_q, \hat{p}_q, \mathbf{X}) = \sum_{m=1}^{M} W_m \big[\sum_{t=1}^{T} \sum_{l=1}^{L} \sum_{k=1}^{K} A_{mtlqk} \\ \cdot W'_m x^{lt} (\phi_{l}(\hat{p}_q) + \Delta p_{mtlqk})\big].
    \end{split}
\end{equation}

where m indexes the attention head, $t$ indexes the frame sampled from the same video clip, and $l$ indexes multi-scale feature maps from the backbone,  and $k$ indexes the sampling points, and $\Delta p_{mtlqk}$ and $A_{mtlqk}$ indicate the sampling offset and attention weights of the $k^{th}$ sampling point in the $l^{th}$ feature map of $t^{th}$ frame and the $m^{th}$ attention head, respectively. Each reference point, represented by normalized coordinates $\hat{p}_q \in [0, 1]^2$, is rescaled using $\phi{l}$ to enable sampling across feature maps $l$ that vary in resolution. The scalar attention weight, denoted by $A_{mtlqk}$, is normalized such that ${\sum_{t=1}^{T} \sum_{l=1}^{L} \sum_{k=1}^{K} A_{mtlqk} = 1}$. The temporal multi-scale deformable attention samples $TLK$ points from $TL$ feature maps instead of $K$ points from single-frame feature maps.

\subsection{Spatio-Temporal Pose Decoder}

In contrast to the spatial pose decoder, the goal of the Spatio-Temporal Pose Decoder (STPD) is to take the temporal feature memory of Spatio-Temporal Deformable Memory Encoder and the temporal pose query of Spatio-Temporal Pose Encoder as inputs, and then output the decoded results of the pose query of frames that are updated by the temporal cues.

\begin{figure}[h]
\centering
\includegraphics[width=0.25\textwidth]{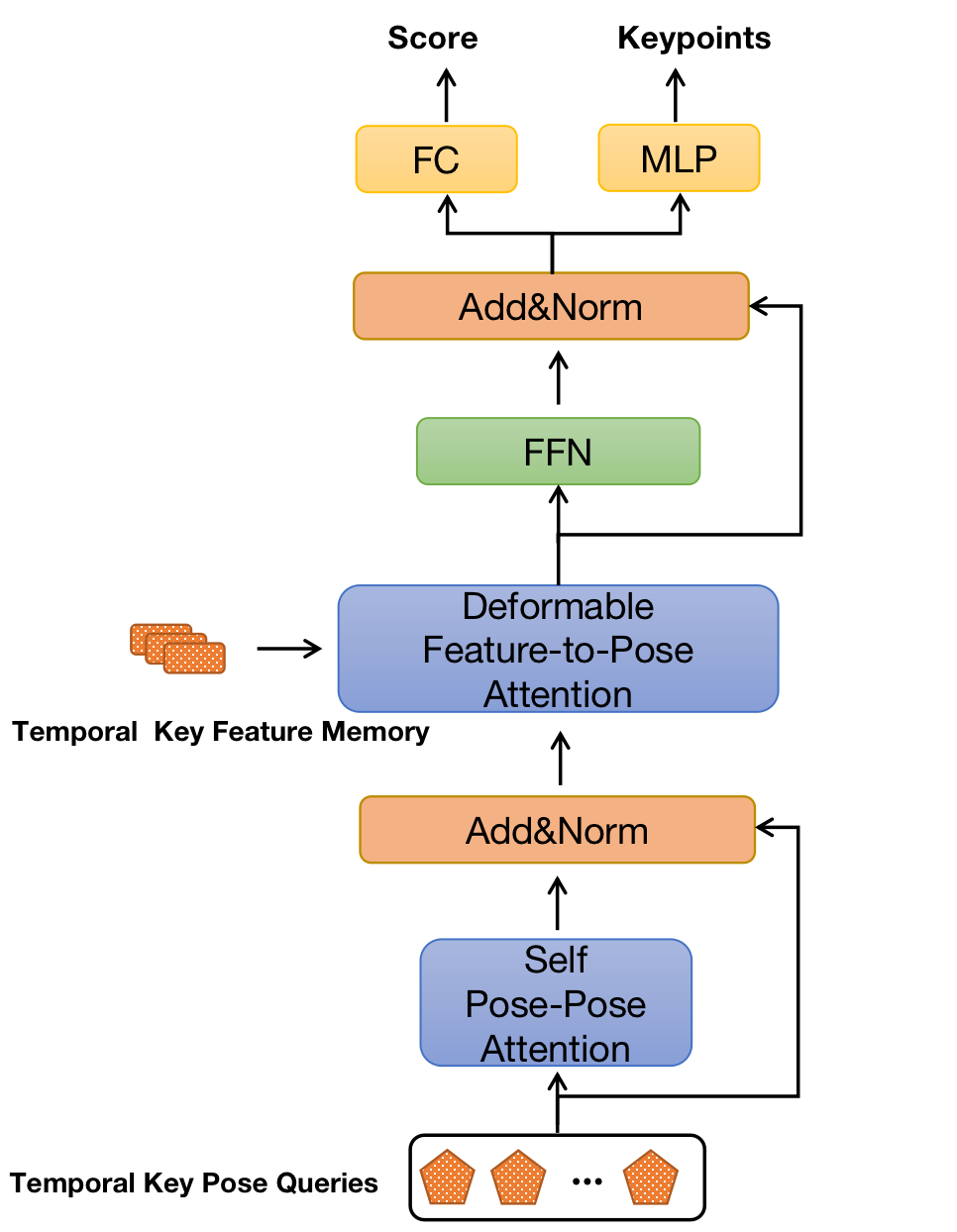} 

\caption{Detailed Architecture of STPD. STPD employs a self-attention module to capture the relationship between spatial objects and cross-attention to capture the interaction between temporal pose query and temporal feature memory.A deformable cross-attention module is designed to focus on visual features closely associated with the target keypoints, and, finally, two separate heads decode the confidence scores and keypoint coordinates, respectively.
}
\label{STPD}
\end{figure}

The implementation is shown in Figure.\ref{STPD}. To simplify the explanation of the process in the figure, we designate one frame as the keyframe.
First, temporal key pose queries from STPE are fed into Self Pose-Pose Attention to capture the relationship between N instances in the frame, which can provide context information between these objects. After Self Pose-Pose Attention, the output query can be represented as follows:
\begin{equation}
\hat{\mathcal{Q}}^t_{pose}=MHA(\bar{\mathcal{Q}}^t_{pose},\bar{\mathcal{Q}}^t_{pose},\bar{\mathcal{Q}}^t_{pose})+\bar{\mathcal{Q}}^t_{pose}
\end{equation}
where HMA stands for the multi-head attention mechanism and $t$ stands for the keyframe.

Then the output $\hat{\mathcal{Q}}^t_{pose}$ as Query interacts with the temporal key multi-scale feature memory  $\bar{\mathcal{F}}^t$ from STDME with the cross-attention mechanism to extract the temporal image features and aggregate them on $\tilde{\mathcal{Q}}^t_{pose}$, and outputs $\tilde{\mathcal{Q}}^t_{pose}$. The process can be expressed as: 
\begin{equation}
\tilde{\mathcal{Q}}^t_{pose}=MSDA(\hat{\mathcal{Q}}^t_{pose},\bar{\mathcal{F}}^t,\bar{\mathcal{F}}^t)+\hat{\mathcal{Q}}^t_{pose}
\end{equation}
where MSDA is multi-scale deformable attention, and $\bar{\mathcal{F}}^t$ is temporal multi-scale feature memroy. 

It is worth noting that the cross-attention uses a multi-scale deformable attention module, which we name Deformable Feature-to-Pose Attention. After $\tilde{\mathcal{Q}}^t_{pose}$ passes through the cross-attention mechanism, it passes through the FFN network, which performs nonlinear transformations and pattern extractions, thus allowing the model to perform deeper representation learning. Finally, the classiﬁcation head predicts the pose conﬁdence score for each query by a linear projection layer.

In the STPD approach, three decoder layers are utilized in a step-by-step manner. Unlike the conventional method of using only the final decoder layer to determine pose coordinates, our model employs all decoder layers for gradual refinement of these coordinates. Specifically, each successive layer reinterprets the pose according to the predictions generated by the preceding layer. Formally, if $Q_{d-1}$ represents the normalized pose predicted by the $(d-1)^{th}$ decoder layer, then the $d^{th}$ decoder layer further refines this pose as follows:

\begin{equation}
\mathcal{Q} _d=  \sigma ( \sigma ^ {-1} (\mathcal{Q}_{d-1})+\Delta \mathcal{Q}_{p})
\end{equation}

In the $d^{th}$ layer, the offsets are predicted as $\Delta \mathcal{Q}_{p}$, where $\sigma$ represents the sigmoid function, and $\sigma ^ {-1}$ denotes its inverse, the inverse sigmoid function. This progressive output prediction with coarseness to fineness can effectively resolve regressions of fine coordinates to prevent misalignment.

\subsection{Instance Consistency Mechanism}

In scenes where subtle human differences in appearance exist, matching pose instances across frames can be challenging due to the highly similar features among different individuals. This similarity can cause errors in matching, which hinders the learning efficiency of the spatio-temporal pose encoder.

To address this issue, we introduced a new learnable embedding  (instance query) to guide pose instances in complex environments for cross-frame matching, thereby improving the performance of the Spatio-Temporal Pose Encoder. Specifically, we created learnable embeddings with the same quantity and feature dimension as the pose queries, where each instance query is linked one-to-one with its corresponding pose query. Using a self-attention mechanism, the instance query interacts with pose queries and visual features to extract and capture each instance's most distinguishing visual characteristics, achieving cross-frame instance tracking through similarity matching. Ultimately, the instance query generates an instance mask by calculating similarity, effectively filtering out irrelevant instances to identify the target that best matches across each frame. This process enhances cross-frame pose matching accuracy and significantly improves the learning efficacy of the Spatio-Temporal Pose Encoder.

At the same time, we introduce the Instance Consistency Mechanism (ICM), designed to mitigate mismatches among  instances through the implementation of Instance Consistency Loss (ICL). The essence of this mechanism is to ensure that instances of the same individual across different frames maintain consistent feature representations. Simultaneously, it aims to increase the spatial feature distance between instances of different individuals, thereby enhancing their distinguishability. As depicted in Fig.\ref{framework}, the Instance Consistency Mechanism is crucial for differentiating between instances. Our method begins with the Hungarian algorithm\cite{kuhn1955hungarian} to establish a one-to-one correspondence between the predicted results and the ground truth (GT) data, thus forming a set of candidate results. We use trackId information to identify  instance query of the same individual across multiple frames as positive samples, and instance query of other individuals as negative samples, using a designated instance as the anchor. Following this, the Instance Consistency Loss (ICL) is introduced to supervise the learning within this candidate set. The formula for instance consistency loss, denoted as

\begin{equation}L_{ic}=\sum_{i=1}^N\left[\max\left(0,d(a_i,p_i)-d(a_i,n_i)+margin\right)\right]\end{equation}

Where $N$ is the number of instances in the candidate set, $a_{i}$ denotes the anchor, i.e., the $i^{th}$ instance, $p_{i}$ denotes the positive sample, i.e., the same individual's pose instance object in other frames, and $n_{i}$ denotes the negative sample, i.e., the different individual's pose instance object. $d(x,y)$ denotes the distance between $x$ and $y$, and cosine similarity is used as the metric distance here. The parameter Margin plays a crucial role, as it regulates the model's ability to effectively discriminate between positive and negative sample pairs. In this way, the instance consistency loss aims to optimize the model performance by ensuring more accurate identification and tracking of the same instance in consecutive frames and reducing the number of mismatch events.

Finally, with the above technique, we successfully realize instance query with high differentiation and one-to-one correspondence with pose query.

\section{Experiments}

\subsection{Datasets}

The PoseTrack dataset represents an extensive collection for human pose tracking and estimation across video sequences. It captures demanding scenarios, incorporating intricate motions and densely populated scenes. In the 2017 version of PoseTrack\cite{iqbal2017posetrack}, there are 514 video segments, containing a total of 16,219 pose labels, split into 250 clips for training, 50 for validation, and 214 for testing. Later, PoseTrack2018\cite{andriluka2018posetrack} augmented this dataset, amassing 1,138 video segments and 153,615 pose annotations, allocated into 593 training clips, 170 validation clips, and 375 testing clips. Both datasets annotate 15 distinct joints, along with additional tags indicating joint visibility. PoseTrack21\cite{doering2022posetrack21} extends the annotations, particularly for small individuals and crowded scenes, with 177,164 pose annotations.  In our model evaluation, we focus on visible joints and use average precision (AP).

\subsection{Experimental setup}

\subsubsection {Training details.} 
Based on prior research, we adopt HRNet-W48\cite{sun2019deep} as the backbone network in our approach. We first performed pre-training on the MS COCO dataset, and the second phase performed 20 epochs of temporal phase training on the Posetrack dataset. Similar to PETR\cite{shi2022end}, we employ the AdamW optimizer with an initial learning rate of $2\times 10^{-4}$ for Transformers and weight decay is set to $10^{-4}$. We initialize the number of pose queries as 100. Throughout training, we use a batch size of 8, and to ensure comparability with prior models, we maintain a fixed number of 3 frames in all experiments. Our training and testing processes are conducted on 8 Tesla V100 GPUs.

\subsubsection{Testing details.} 
The input images are adjusted to ensure their shortest edges measure 800, while their longest edges do not exceed 1333. The evaluation time is recorded with a single NVIDIA Tesla V100 GPU.

\subsubsection{details on inference speed}
The data is presented in Table.\ref{17val}  illustrates the average inference durations across the PoseTrack2017 dataset for various models, calculated per frame. On average, each frame of the PoseTrack2017 dataset contains six human instances. The inference times for all evaluated methods were tested using a V100 GPU. Notably, the top-down model employs the YOLOv3\cite{redmon2018yolov3} model as its human detector.

\renewcommand\arraystretch{1.2}
\begin{table*}[h]
    \caption{Quantitative outcomes on the PoseTrack2017 validation set. Note that all of the above except VEPE are multi-stage models. The time indicated represents the model's inference duration across the entire PoseTrack dataset, averaged per frame. The inference times for all methods were tested on a V100, with the top-down models utilizing the YOLOv3\cite{redmon2018yolov3} model as the human detector.
} 
  \resizebox{1\textwidth}{!}{

  \begin{tabular}{lccccccccc}
    \hline
      Method                            &Shoulder &Head    &Elbow       &Wrist   &Hip    &Ankle  &Knee      &{\bf Mean} & {\bf Time [ms]}\cr 
      \hline
      PoseTracker \cite{girdhar2018detect}   &$70.2$  &$67.5$    &$62.0$      &$51.7$  &$60.7$ &$49.8$ &$58.7$         &{$60.6$}     &-\cr
     PoseFlow \cite{xiu2018pose}             & $73.3$ &$66.7$   &$68.3$      &$61.1$  &$67.5$ &$61.3$ &$67.0$        &{$ 66.5$}      &-\cr
JointFlow \cite{doering2018joint}            & -     & -       &-           &-       &-      &-      &-            &{ $ 69.3$}      &-\cr
   FastPose \cite{zhang2019fastpose}   	     &$80.3$ &$80.0$    &$69.5$      &$59.1$  &$71.4$ &$59.4$  &$67.5$        &{$ 70.3$}      &-\cr
   TML++ \cite{hwang2019pose}    	 		 &-       &-     &-      &-      &-      &-       &-    &{$ 71.5$}                  &-\cr
Simple (ResNet-50) \cite{xiao2018simple}     &$80.5$ &$79.1$    &$75.5$      &$66.0$  &$70.8$ &$61.7$  &$70.0$       &{$72.4$}         &-\cr
Simple (ResNet-152) \cite{xiao2018simple}    &$83.4$ &$81.7$    &$80.0$      &$72.4$  &$75.3$ &$67.1$ &$74.8$       &{$ 76.7$}     &{$560$}\cr
  STEmbedding \cite{jin2019multi}            &$81.6$ &$83.8$    &$77.1$      &$70.0$  &$77.4$ &$70.8$ &$74.5$       &{$ 77.0$}       &-\cr
        HRNet \cite{sun2019deep}             &$83.6$ &$82.1$    &$80.4$      &$73.3$  &$75.5$ &$68.5$ &$75.3$       &{$ 77.3$}      & {$580$}\cr
         MDPN \cite{guo2018multi}            &$88.5$ &$85.2$    &$83.9$      &$77.5$  & $79.0$ &$71.4$ &$77.0$        &{$ 80.7$}      &-\cr
   Dynamic-GNN \cite{yang2021learning} 	     &$88.4$ &$88.4$    &$82.0$      &$ 74.5$ &$79.1$ &$73.1$ &$78.3$        &{ $81.1$}         &-\cr
 PoseWarper \cite{bertasius2019learning}     &$88.3$ &$81.4$    &$83.9$      &$ 78.0$ &$82.4$ &$73.6$ &$80.5$        &$ 81.2$         & $1293$\cr
   DCPose \cite{liu2021deep}                 &$ 88.7$ &$ 88.0$       &$ 84.1$   &$78.4$&$ 83.0$        &$ 74.2$ &$ 81.4$       &$ 82.8$      & $1390$\cr
   FAMI-Pose \cite{liu2022temporal}          &$ 90.1$ &$ 89.6$   &$ 86.3$ &$ 80.0$ &$ 84.6$ &$ 77.0$  &$ 83.4$            &$ 84.8$     & $1577$\cr
	
     \bf VEPE (Ours)	      &$\bf 88.5$ &$\bf 87.6$  &$\bf 84.5$ &$\bf 78.8$ &$\bf 83.3$ &$\bf 74.7$ &$\bf 81.7$   &$\bf 83.0$       & $334$ \cr
    \hline
    
    \end{tabular}
    }
    \label{17val}
\end{table*}

\subsection{Comparison with State-of-the-art Approaches
}
\subsubsection{Results on the PoseTrack2017 Dataset.}  

We initially test our model on the PoseTrack2017 validation dataset. In total, $15$ different methods are compared, including PoseTracker \cite{girdhar2018detect}, PoseFlow \cite{xiu2018pose}, JointFlow \cite{doering2018joint}, FastPose \cite{zhang2019fastpose}, TML++ \cite{hwang2019pose}, and SimpleBaseline (both ResNet-50 and ResNet-152). Additionally, comparisons are made with STEmbedding \cite{jin2019multi}, HRNet \cite{sun2019deep}, MDPN \cite{guo2018multi}, Dynamic-GNN \cite{yang2021learning}, PoseWarper \cite{bertasius2019learning}, DCPose \cite{liu2021deep}, FAMI-Pose\cite{liu2022temporal}, as well as our VEPE model. Table \ref{17val} provides the results of these methods on the PoseTrack2017 validation set.

From the table, we can observe that our method outperforms most of the two-stage models and slightly outperforms the DCPose model, and our inference speed substantially outperforms the DCPose and FAMIPose models. In addition, we obtain excellent accuracy at challenging joints such as ankles, wrists, knees, etc., which indicates that our model is robust to fast-motion and body-obscuring scenarios and has excellent inference efficiency.

\begin{figure}[h]
\centering
\includegraphics[width=0.45\textwidth]{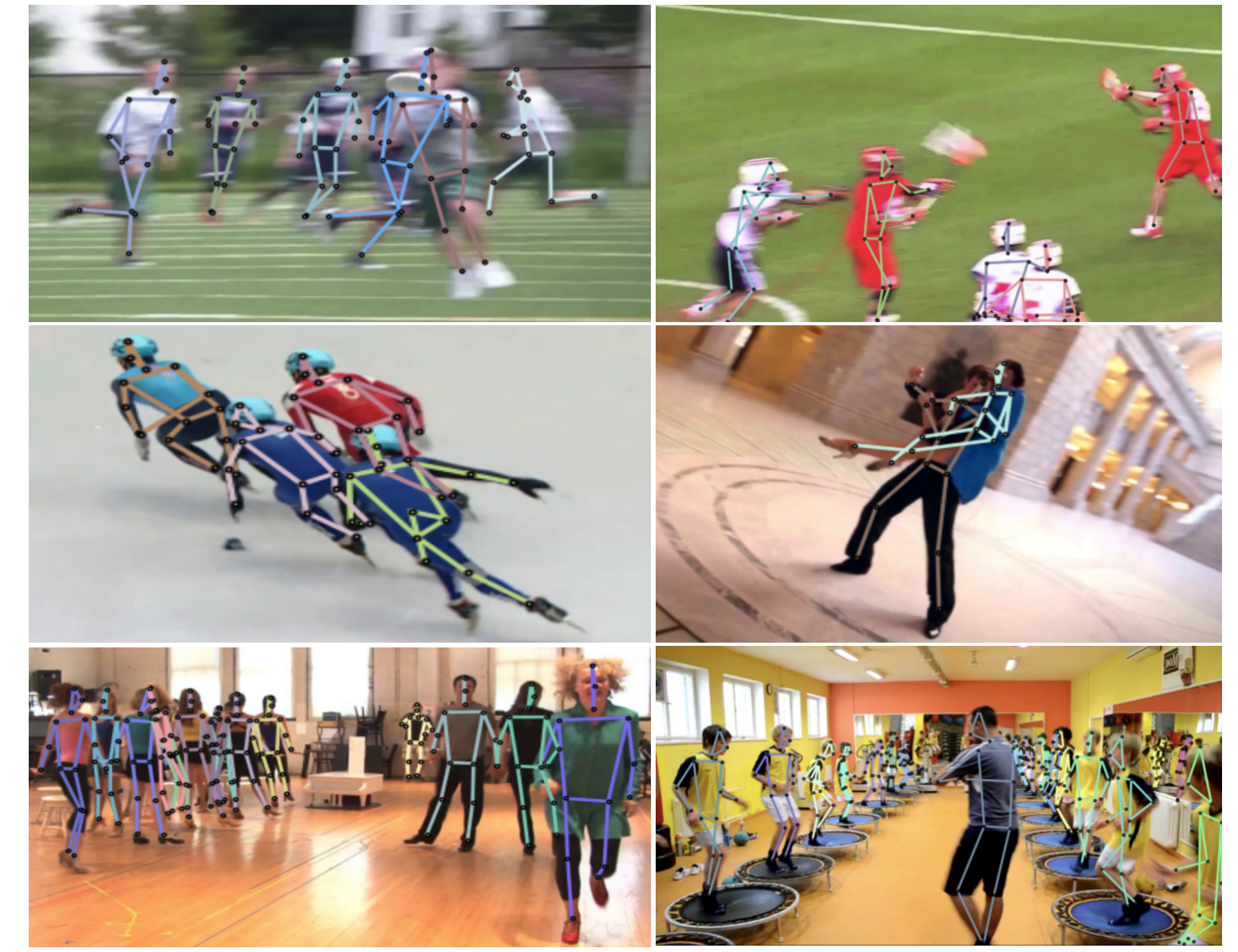} 

\caption{Visualization of Posetrack21 validation results. The first row shows the scene in fast motion, the row column shows the scene where the human body is occluded, and the third row shows the scene with a dense crowd.}
\label{fig:sample}
\end{figure}

\subsubsection{Results on the PoseTrack2018 Dataset.}

We conducted an evaluation of our model using the PoseTrack2018 validation set, making comparisons across a total of 10 different methods. The methods included STAF\cite{raaj2019efficient}, PGPT\cite{bao2020pose}, AlphaPose\cite{fang2017rmpe}, TML++ \cite{hwang2019pose},Dynamic-GNN\cite{yang2021learning}, MDPN \cite{guo2018multi}, PoseWarper\cite{bertasius2019learning}, DCPose\cite{liu2021deep}, FAMIPose\cite{liu2022temporal}, as well as our proposed VEPE. From the table \ref{18val}, VEPE has a simple and flexible end-to-end structure that outperforms most of the two-stage models and improves 0.2 mAP over DCPose.

\subsubsection{Results on the PoseTrack2021 Dataset.}
Our model is evaluated using the validation set from PoseTrack2021. A total of 7 methods are compared in the table \ref{21val}, including Tracktor++ w. poses\cite{bergmann2019tracking,doering2022posetrack21}, CorrTrack\cite{rafi2020self,doering2022posetrack21}, CorrTrack w. ReID\cite{rafi2020self,doering2022posetrack21}, Tracktor++ w. Corr.\cite{bergmann2019tracking,doering2022posetrack21} , DCPose\cite{liu2021deep}, FAMIPose\cite{liu2022temporal} and our VEPE. 

The difference between PoseTrack21 and PoseTrack2018 is the addition of more challenging pose test scenarios such as small figures and figures in crowds. From the table, we can find that in the more challenging PoseTrack21, our model has nearly improved compared to DCpose by 0.3mAP, which shows that our model still has good robustness in complex scenes.
Specific visualization examples are shown in Figure.\ref{fig:sample}

\subsection{Ablation Experiments}
We conducted ablation studies focusing on the contribution of each component in the VEPE framework. We also examined the STPE and Instance Consistency Mechanism for further discussion. All experiments were performed on the PoseTrack 2017 validation set.

\subsubsection{Study on components of VEPE}

Table.\ref{ab-1} presents a summary of the impact of various design components on the posetrack17 dataset. The baseline model refers to the model with only the spatial phase of training, which has a mAP of 77.2. Finally, all components were used to reach a mAP of 83.0, which is an improvement of 5.8 mAP with respect to the baseline model, with the Spatio-Temporal Pose Encoder (STPE) module showing the largest improvement.

\begin{table}[h]
  \centering
\caption{\centering Ablation of different components in VEPE.}
  \resizebox{0.45\textwidth}{!}{
  \begin{tabular}{c|cccc|c}
    \hline
     Method  &STPE & ICM &STDME &STPD   &Mean\cr
    \hline
    Baseline & & & & &77.2 \cr
    (a) &\checkmark & & &  &$80.7$ \cr
    (b) &\checkmark &\checkmark & &  &$82.1$\cr
    (c) &\checkmark & \checkmark &\checkmark &  &$\bf82.5$\cr
    
    (d) &\checkmark & \checkmark &\checkmark & \checkmark & $\bf83.0$\cr

    \hline
    \end{tabular}}
    \label{ab-1}

\end{table}

\subsubsection{Study on Pose Query Selection}
In dense scenes, instance frequently exhibit considerable overlap and similarity due to the static characteristics common to human bodies. To tackle this redundancy and improve the efficacy of the spatio-temporal pose encoder (STPE), we have developed a module specifically for selecting pose queries. This module enhances query quality by employing a confidence threshold, appropriately set by the pose instance evaluator, to filter out queries. The optimal threshold is established through rigorous experimental analysis. The effects of different confidence thresholds on the selection of pose queries are detailed in the following table\ref{tab:PQS}.

\begin{table}[h]
\centering
    \caption{\centering Effect of Different Threshold Settings on Pose Query Selection}

\begin{tabularx}{0.45\textwidth}{>{\centering\arraybackslash}m{1.5cm}|*{5}{>{\centering\arraybackslash}X}}        \toprule
        threshold & 0.1& 0.2 & 0.3 & 0.4 & 0.5  \\
        \midrule
        mAP(\%) & 80.4 & 80.5 & 80.7 & 80.4 & 80.1  \\
        \bottomrule
    \end{tabularx}
\label{tab:PQS}
\end{table}


%

\subsubsection{Study on Instance Consistency Mechanism}

The instance consistency mechanism aims to improve the accuracy of the temporal pose encoder in pose query recognition and tracking and reduce the mismatch problem in cross-frame pose query matching by enhancing cross-frame instance query consistency and differentiation. Briefly, we introduce instance queries that correspond one-to-one with the pose query and ensure that the instance query feature space representations of the same instance in different frames have similarity. The tracking of instances is realized by calculating the similarity between instance queries, which guides the pose query for cross-frame matching and improves the accuracy of the temporal pose encoder in instance identification and tracking.

\begin{figure}[h]
\centering
\includegraphics[width=0.5\textwidth]{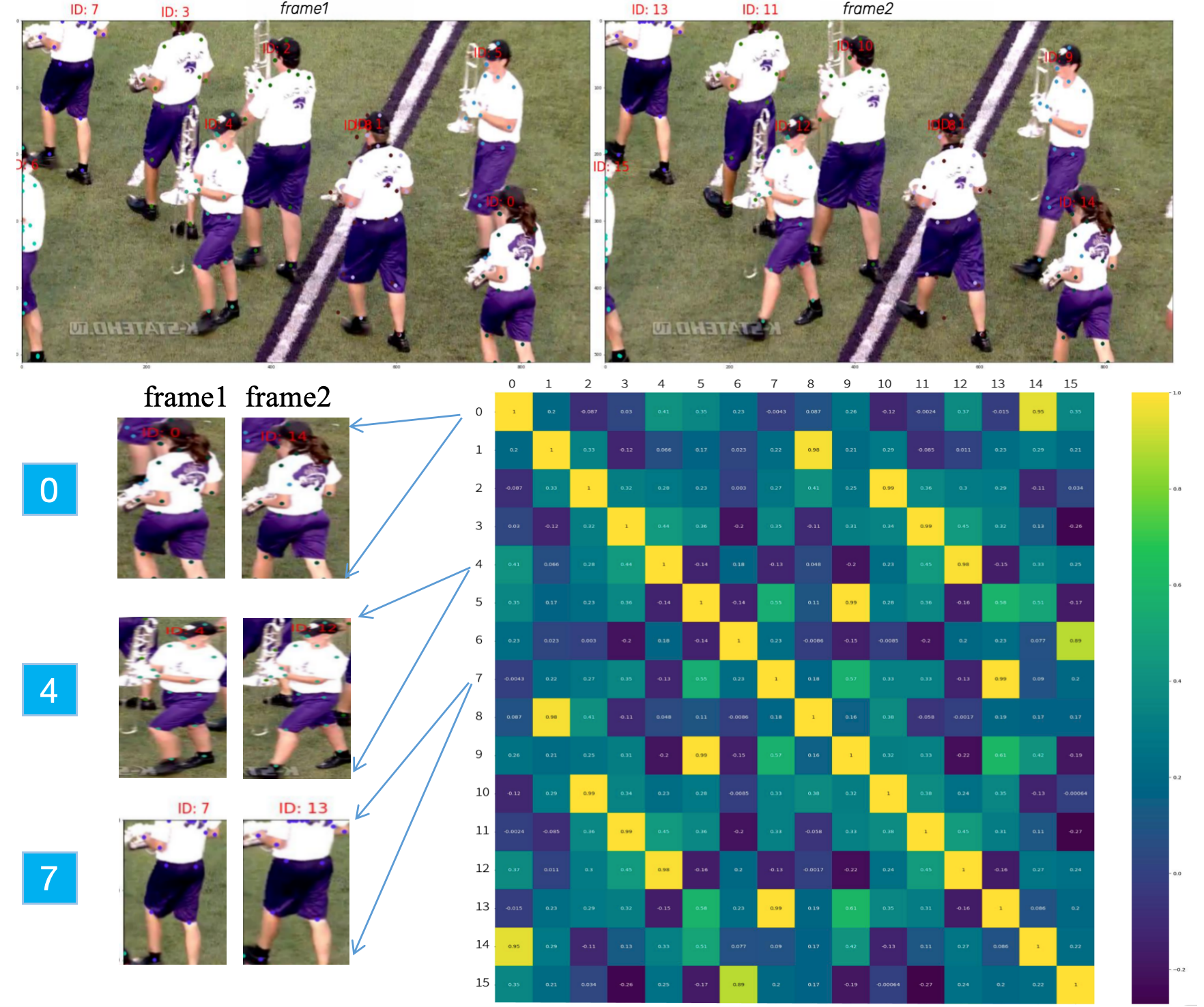} 

\caption{\centering Instance tracking between cross-frame instance queries}
\label{fig:similar}
\end{figure}

As can be seen from Table.\ref{ab-1}, the inclusion of the Instance Consistency Mechanism (ICM) improves the model accuracy by 1.4\%, which proves the effectiveness of this mechanism. Figure.\ref{fig:similar}  illustrates the tracking effect of instance queries under the instance consistency mechanism, using two example frames. The heatmaps in the lower-right corner show the similarity between all instance queries, allowing for the identification of the same individual across frames. For instance, the similarity heatmap in the first row shows that ID-0 and ID-14 represent the same human individual and they have the highest similarity.

\section{Conclusion}

In this paper, we propose a simple video-based end-to-end multi-person pose estimation framework, VEPE, which uses the Transformer to naturally associate instance targets in a video as a sequence-to-sequence task and performs end-to-end training without any post-processing. The VEPE  learns the temporal context of pose instances and visual cues across frames through the Spatio-Temporal Pose Encoder (STPE) and the Spatio-Temporal Deformable Memory Encoder (STDME), and finally, Spatio-Temporal Pose Decoder (STPD) cascades to update the pose results of the keyframes by utilizing the spatio-temporal cues from Spatio-Temporal Pose Encoder and Spatio-Temporal Deformable Memory Encoder. Furthermore,  to reduce the mismatch problem during the cross-frame pose query matching process, an instance consistency mechanism is introduced, which aims to enhance the consistency and discrepancy of the cross-frame instance query and realize the instance tracking function, which in turn guides the pose query to perform accurate cross-frame matching. Extensive experiments on the Posetrack dataset demonstrate that VEPE surpasses most two-stage models in performance and triples inference efficiency.

{
    \small
    \bibliographystyle{ieeenat_fullname}
    \bibliography{main}
}


\end{document}